\documentclass[letterpaper, 10 pt, conference]{ieeeconf}  % Comment this line out if you need a4paper

\IEEEoverridecommandlockouts                              % This command is only needed if 
\usepackage{graphics} % for pdf, bitmapped graphics files
\usepackage{epsfig} % for postscript graphics files
\usepackage{mathptmx} % assumes new font selection scheme installed
\usepackage{times} % assumes new font selection scheme installed
\usepackage{amsmath} % assumes amsmath package installed
\usepackage{amssymb}  % assumes amsmath package installed

\usepackage{color}
\usepackage{xcolor}
\usepackage{caption}
\usepackage{preambles}

\title{\LARGE \bf
Reciprocal Collision Avoidance for General \\Nonlinear Agents using Reinforcement Learning}

\author{{Hao Li\textsuperscript{1}, Bowen Weng\textsuperscript{1}, Abhishek Gupta\textsuperscript{1}, Jia Pan\textsuperscript{2}, and Wei Zhang\textsuperscript{3}}
\thanks{\textsuperscript{1}Hao Li, Bowen Weng, Abhishek Gupta are with the Department of Electrical and Computer Engineering, The Ohio State University, Columbus, OH, USA. (e-mail: \{li.7857, weng.172, gupta.706\}@osu.edu)
}
\thanks{\textsuperscript{2}Jia Pan is with the Department of Computer Science, The University of Hong Kong, Hong Kong. (email: panjia1983@gmail.com)
}
\thanks{\textsuperscript{3}Wei Zhang is with SUSTech Institute of Robotics, Southern University of Science and Technology, China. (email: zhangw3@sustech.edu.cn)
}
}

\usepackage[noadjust]{cite}
\usepackage{caption}
\usepackage{subcaption}
\usepackage{algorithm,algpseudocode}
\def\R{\mathbb{R}}
\def\Na{\mathbb{N}}
\def\s{\mathbf{s}}
\def\a{\mathbf{a}}
\def\o{\mathbf{o}}
\def\p{\mathbf{p}}
\def\y{\mathbf{y}}
\def\w{\mathbf{w}}
\def\ve{\mathbf{v}}
\def\S{\mathcal{S}}
\def\A{\mathcal{A}}

\begin{document}
\maketitle
\thispagestyle{empty}
\pagestyle{empty}

\begin{abstract}
Finding feasible and collision-free paths for multiple nonlinear agents is challenging in the decentralized scenarios due to limited information of other agents and complex dynamics constraints. In this paper, we propose a fast multi-agent collision avoidance algorithm for general nonlinear agents with continuous action space, where each agent observes only positions and velocities of nearby agents. To reduce online computation, we first decompose the multi-agent scenario and solve a two agents collision avoidance problem using reinforcement learning (RL). When extending the trained policy to a multi-agent problem, safety is ensured by introducing linear constraints from the optimal reciprocal collision avoidance (ORCA) and the overall collision avoidance action could be found through a simple convex optimization. Most existing RL-based multi-agent collision avoidance algorithms rely on the direct control of agent velocities. In sharp contrasts, our approach is applicable to general nonlinear agents. Realistic simulations based on nonlinear bicycle agent models are performed with various challenging scenarios, indicating a competitive performance of the proposed method in avoiding collisions, congestion and deadlock with smooth trajectories. 

\end{abstract}

% Advantages: Nonlinear, partial observation, Safety, Deadlock, Easy to train(complexity grows linearly), limited range of sensing

\section{Introduction}
Multi-agent navigation is an important topic in robotics with many potential real-world applications such as service robots, logistic robotics, search and rescue, and self-driving car, among many others. One of the main challenges in designing navigation algorithms is on the design of a real-time control policy that satisfies safety requirements while respecting the dynamics of each agent.

General multi-agent navigation algorithms can be categorized into centralized algorithms and decentralized ones. The centralized algorithm considers all agents and the environment as a whole system~\cite{sanchez2002using,vsvestka1998coordinated}. A centralized decision maker is in charge of the monitoring and control of all the agents in the system. In practice, such a centralized approach may encounter many issues. First, this method heavily relies on fast communication and computation, hence is very sensitive to signal delay, which is common in real-world implementations. Second, whenever the central agent fails, all the agents would lose control and may lead to collisions. Finally, the centralized algorithm often scales poorly to large-scale systems.

Contrary to centralized approaches, decentralized algorithms allow each agent to make decision independently by taking into account the limited information from sensors and communications. Similar to many studies in the literature, in this paper, we consider the realistic case where there is no real-time communication among agents. This directly leads to one obvious challenge, namely, prediction of the intents and actions of other agents. One of the most popular methods to handle this problem is the so-called optimal reciprocal collision avoidance (ORCA)~\cite{van2011reciprocal}. Instead of explicitly making predictions regarding other agents, ORCA provides a safety guarantee by computing linear velocity constraints for each agent. It is based on the idea of velocity obstacles (VO)~\cite{van2008reciprocal,fiorini1998motion}. By introducing a pair of linear constraints in the velocity space for each pair of nearby agents, the ORCA method transforms the collision avoidance problem into a linear programming problem that can be solved efficiently. Notice that the original ORCA method assumes the agent model is a simple single integrator, where the velocity of agents can be directly controlled and changed instantaneously, which may not be realistic for real robots. The ORCA method has been extended to more general cases, including second-order system with acceleration control~\cite{van2011reciprocal2}, dynamics with continuity constraints~\cite{rufli2013reciprocal}, differential drive robots~\cite{alonso2013optimal}, car-like robots~\cite{alonso2012reciprocal} and heterogeneous system of various types of agent dynamics~\cite{bareiss2015generalized}. 
Most of such works are designed to certain specific dynamics and controllers. They transfer the control of agents to the velocity space by using fixed tracking controllers, which take velocity reference as input. Although the method in~\cite{bareiss2015generalized} can be applied to general nonlinear systems, it assumes that each agent could observe actions of the other agents, which confines the application to agents with very simple dynamics.
% r. They transfer the control of agents to the velocity space, which is confined to certain specific dynamics and controller. 
In addition, all ORCA-based methods require preferred velocities or control from a high-level planning module. If the planning is not well designed, the generated trajectories could be very unnatural, and sometimes leads to deadlock behaviors~\cite{godoy2018alan}.

Another approach to handle the multi-agent navigation is using reinforcement learning (RL)~\cite{chen2017decentralized, chen2017socially, everett2018motion, fan2018fully, long2018towards}. While it has been demonstrated that RL can solve the optimal control problem of complex nonlinear dynamics~\cite{lillicrap2015continuous}, the problem nature of multi-agent navigation has presented several key challenges. First, the number of agents involved in the planning period is dynamically
varying, and could be different from the number of agents used for training. To address this problem, the authors in~\cite{chen2017decentralized} proposed to train a value function with two agents, and then extend the obtained policy to the multi-agent case by selecting the action that maximizes the minimum of pairwise value functions. Symmetric network is introduced in~\cite{chen2017socially} and long short-term memory (LSTM) neural network is used to handle this problem~\cite{everett2018motion}. The laser scanner data is also adopted as observation state in~\cite{fan2018fully} and~\cite{long2018towards}. However, most of the aforementioned studies are restricted to discrete action space, and require rather complex neural networks with demanding training process. Besides, they all assume that velocities of robots could be directly controlled, which limits their applications to robots with more complex nonlinear dynamics. Another key issue of in RL-based approaches is the difficulty in providing safety guarantees. The performance is often limited to scenarios encountered during training. In fact, to the best of our knowledge, most existing RL-based multi-agent collision avoidance methods do not systematically incorporate safety.

% This algorithm is only applicable to discrete action space and may suffer from the instability of value-function based algorithm.  However, they are both restricted to discrete action space with large network and complex training procedure.
% have large network with complex training procedure, and discrete action space. For \cite{fan2018fully} and \cite{long2018towards}, they use laser scanner data as reinforcement learning state, so they do not have the problem. But their training procedure is very complex and has large neural network for training. 
% So far all learning based approach has only been demonstrated with very simple dynamics, and there is no safety guarantee. 
% On the contrary, our algorithm could be applied to general nonlinear system with safety guarantee. We only need to do the training for two agents, and we could use nearly all reinforcement learning algorithm to do the training. Our network structure is relatively small and the training procedure is really fast, which has been demonstrated in our simulation.
% There are also some commonly used distributed approaches like potential field methods, probabilistic road-maps and decentralized MPC. But in general they could not solve deadlock and need heuristic method to alleviate the problem. 

In this paper, we propose an efficient RL-based algorithm to solve multi-agent collision avoidance problem with a systematic consideration of safety of the overall system. Our approach is designed for robots with general nonlinear dynamics, where each agent can only observe positions and velocities of nearby robots. We first decompose our problem into a two agents collision avoidance problem with continuous action space. Then we handle the multi-agent collision avoidance problem by solving a simple convex optimization with safety constraints from ORCA. The main contributions of this work are summarized below:
\begin{list}{$\bullet$}{\topsep=0.ex \leftmargin=0.15in \rightmargin=0.in \itemsep =0.02in}
\item \textbf{General Nonlinear Agents:} Unlike other RL-based approaches that assume velocities of robots could be directly controlled, or other ORCA-based approaches that need specific models and controllers, our approach works for general nonlinear systems with continuous action space. 
% For example, in the bicycle model used Section~\ref{simuresults}, the control inputs are the steering angle and forward acceleration.
\item \textbf{Improved Safety:} Different from other RL-based approaches which only consider safety in reward function design, our approach incorporates safety bounds systematically by introducing linear constraints obtained from ORCA. In Section~\ref{simuresults}, we demonstrate the safety of our algorithm with complex and challenging tasks.
\item \textbf{Lightweight Structure:} Through decomposing the problem into a two agents collision avoidance problem, we significantly simplify the RL training task. The learning process of our approach is much faster and the neural network is much smaller when compared with other RL-based approaches.
\end{list}

% \noindent \textbf{Main results:} In this paper, we address the decentralized multi-agent collision avoidance problem by proposing a safe and fast algorithm, which could be applied to robots with general nonlinear dynamics by only observing the position and velocity of nearby robots. We first decompose our problem into pairwise collision avoidance problem with continuous action. When comparing with other RL-based approach, the learning process of our approach is really fast since the state space is small for only two agents. Nearly all RL algorithms could be easily embedded in our approach with small neural networks. Different from other RL-based approaches, we introduce a safety guarantee by transferring the multi-agent collision avoidance into an optimization problem with constraints from ORCA. We demonstrate our algorithms with various challenging tasks in simulations, which shows our approach could easily handle collision avoidance, congestion and deadlock with smooth trajectories.

% \section{Background}
% \subsection{Deep Reinforcement Learning}
% \subsection{Optimal Reciprocal Collision Avoidance}

\section{Problem Formulation}
We consider a collision avoidance problem for $M$ nonlinear agents with the following dynamics:
% We consider here the problem of multi-agent collision avoidance with $M$ homogeneous robots. Each agent has the same constrained nonlinear dynamics,
\begin{equation}
\begin{cases}
\s^i_{k+1} = f(\s^i_k, \a^i_k), \label{eq:dynammics} \\
\y^i_{k} = h(\s^i_k),
\end{cases}
\end{equation} 
where $f(\cdot, \cdot)$ is the nonlinear state-transition function, and $h(\cdot)$ is the output function. We denote the state of the $i$-th agent at discrete time $k$ with $\s^i_k\in \S \subset \R^n$, and let $\a_k^i \in \A\subset \R^m$ denote its own action at time $k$, where $i = 1, 2, \ldots,  M$. Here $\S$ and $\A$ represent convex constraints in state space and action space, respectively.
The output of the system is $\y^i_k = [\p^i_k, \ve^i_k]$, which consists of the position $\p^i_k \in \R^d $ and velocity $\ve^i_k \in \R^d $ of agent $i$. For example, if the state $\s^i_k$ comprises the position $\p^i_k $ and the velocity $\ve^i_k$ and some other information, then another agent $j\neq i$ observes only the output of the agent $i$, i.e., the position and the velocity. Given $M$ total agents, for agent $i$, it could observe output of all other agents: $\y^{-i}_k \triangleq [\y^1_k, \y^2_k, \ldots, \y_k^{i-1}, \y_k^{i+1}, \ldots, \y_k^{M} ]$. 
% Each robot knows its current state and goal state, and also observes the position and velocity of other robots (but not the goal state of the other robots).  
% The multi-agent collision avoidance problem is based on the assumption that each robot has the same dynamics, that is, they are homogeneous.  Each agent knows its own state and goal, and could observe nearby agents.

Each agent has a goal state $\tilde{\s}^i$, which is the state they would like to reach. It is assumed that each agent knows its own state and goal state, and observe output of other agents. All the information agent $i$ has at time $k$ could be represented as observation, $\o_k^i\triangleq [\s^i_k, \tilde{\s}^i, \y^{-i}_k]$.
% , which is defined in a global coordinate system. In practice each agent observe other agents from its own configuration, hence we need to transfer the observation into a local coordinate system for each agent, $\tilde{\o}_k^i = g(\bar{\o}_k^i) \triangleq [\tilde{\s}^i_k, \tilde{\s}^i_g, \tilde{\underline{\o}}^i_k]$, here $g(\cdot)$ is the coordinates transformation function from global coordinate systems to local coordinate systems, and $\tilde{\o}_k^i$ is called the \textit{local joint observation}, $\tilde{\s}^i_k$, $\tilde{\s}^i_g$, and $\tilde{\underline{\o}}^i_k$ represents the state, goal state and observation of other agents in the new local coordinate system. Here we use the tilde symbol to represent the corresponding variable in the new local coordinate system. For example, if each agent is a car, then the global coordinate system is fixed with respect to the global positioning system, then the origin of a local coordinate system could be defined as the centroid of the car, with x-axis pointing towards the head of the vehicle along the longitudinal axle. We need to transform the current state, goal state and observation of other vehicles into the new coordinate system. 
Given the observation, we would like to compute a control policy $\pi$ that maps the observation $\o^i_k$ to its action $\a^i_k$ for all robots.
% Note that the policy is not dependent on the robot, since all the robots are homogeneous.
% and the observations are in the local coordinate system). 
We parameterize the policy using variable $\theta\in\R^n$, that is
\begin{align}
\a^i_k = \pi(\o^i_k, \theta) \in \A
\end{align}
% To check collision, it is assumed that each agent could be approximated as a disc with the same radius $r$. 

Given the policy $\pi(\cdot, \theta)$, each agent should reach their goal state in the shortest time while avoiding collision with each other. Then the collision avoidance problem could be formulated as a multi-agent sequential decision making problem, given by
\begin{subequations}
\label{eq:optimization}
\begin{align}
\min_{\theta}\ &\frac{1}{M}\sum_{i=1}^M \sum_{k=0}^\infty \gamma^{k} R(\o_k^i) \\
\text{s.t. } &\s^i_{k+1} = f(\s^i_k, \a^i_k)\ \forall i = 1,\ldots, M, \: \forall k \in\Na \label{cons:dynamics} \\
& \a^i_k = \pi(\o^i_k, \theta) \in \A\label{cons:ctrl}  \\
& \|\p^i_k - \p^j_k \| > 2r, \ \forall i, j = 1,\ldots, M,\: i \neq j,\: \forall k \in\Na\label{cons:col}\\
& \s_0^i = \s^i_{initial}, \ \forall i = 1,\ldots, M, \label{cons:init}
\end{align}
\end{subequations}
where $R(\cdot)$ is a scalar reward function, and $\gamma \in (0, 1]$ is the discount factor. The reward function $R(\cdot)$ awards the robot for approaching the goal state and penalizes for collisions with other agents, it will be further explained in Section~\ref{sub:two}. Equation~(\ref{cons:dynamics}) are due to nonlinear dynamics, equation~(\ref{cons:ctrl}) means that all agents should follow the same policy, and inequalities~(\ref{cons:col}) encode collision avoidance conditions. It is assumed that each agent could be approximated as a disc with the same radius r. Each agent $i$ has a given initial state $\s^i_{initial}$, which is expressed as the equality constraint~(\ref{cons:init}).

It is rather challenging to directly solve~(\ref{eq:optimization}) in a centralized way. The optimization is highly non-convex due to nonlinear equality constraints~(\ref{cons:dynamics}) and non-convex inequality constraints~(\ref{cons:col}). Besides, each agent only has partial information of surrounding agents. Instead of solving~(\ref{eq:optimization}) directly, we handle the problem by first solving a two agents collision avoidance problem via RL, and then transfer the multi-agent collision avoidance problem into a simple convex optimization problem with safety constraints from ORCA. 
% Instead of solving~(\ref{eq:optimization}) directly, we handle the problem in a decentralized approach by first decomposing the problem into pairwise collision avoidance problem. The two agents problem could be solved via deep RL in a short time, and a policy with continuous action space could be returned. Given the trained policy, we extend the solution to the multi-agent problem via a simple convex optimization with ORCA constraints for safety guarantee.

\section{Approach}
\begin{figure}[t]
    \centering
    \includegraphics[trim={0.1cm 3cm 8cm 0.2cm},clip,width=0.55\textwidth]{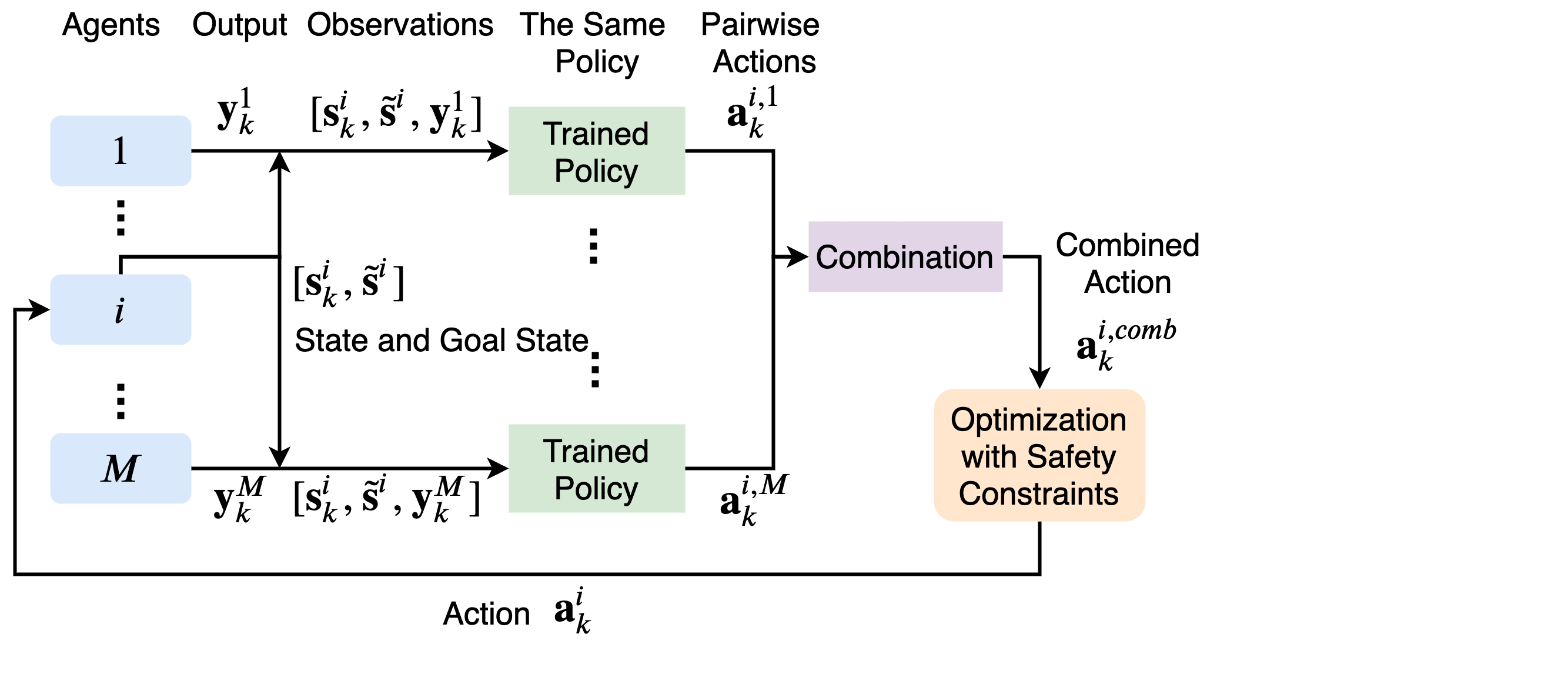}
    \caption{The Algorithm Framework}
    \label{fig:framework}
\end{figure}

This section presents an algorithm to solve the multi-agent collision avoidance problem~(\ref{eq:optimization}) which combines RL and ORCA. To reduce online computation and make the training fast, we first solve a two agents collision avoidance problem using deep RL. 
Many exising RL-based collision avoidance algorithms 
% like~\cite{chen2017decentralized,chen2017socially,everett2018motion}, 
assume that agents can directly control their velocities and require discretization of the action space. Thus, they are only applicable to simple dynamics and can lead to unnatural trajectories. Different from those studies, we formulate RL with continuous action space for general nonlinear agents. Our RL-based approach is faster than other RL-based algorithms with a lightweight policy neural network. Given the trained policy, we apply it to multi-agent problems by combing pairwise actions introduced by different nearby agents. 
% Similar to other RL-based approach, it is possible that the combined action may lead to collisions. 
To avoid collisions, ORCA is introduced and we selects action by solving a simple convex optimization problem with linear safety constraints. The framework of our algorithm is shown in Fig.~\ref{fig:framework}.

\subsection{Two Agents Collision Avoidance via Reinforcement Learning}\label{sub:two}
\textit{\textbf{Network Structure:}} For general policy-based RL algorithm, the policy is usually represented as a neural network, with the state as input and the action or parameters of the action as output. For the two agents collision avoidance problem with agents $i$ and $j$, both of them have their observations. With slight abuse of notation, we denote the observations of agent $i$ and $j$ by $\o^{i,j}_k$ and $\o^{j,i}_k$, respectively. State space in standard RL is now the observation space. We use a deterministic policy, where the input of the neural network is the observation and output is the action, that is $\a^{i,j}_k = \underline{\pi}(\o^{i,j}_k, \underline{\theta} )$. Here we use $\underline{\pi}(\cdot, \cdot)$ to denote the policy neural network and $\underline{\theta}$ to denote parameters of the network. Since there are only two agents, the input dimension of the neural network is relatively small and the training can be done efficiently.
% With a little abuse of notation, the policy of agent $i$ at time $k$ could be represented as $\pi(\bar{\o}^{i,j}_k, \theta)$, here $\theta$ represents weights of the neural network.

\textit{\textbf{Reward Design:}} The objective for each agent is to achieve the goal state in the shortest time while avoiding collision with each other. The reward function is designed as follows:
\begin{align}
\label{eq:rewards}
R(\o^i_k) = (r_g)^i_k + (r_p)^i_k + (r_c)^i_k
\end{align}
For agent $i$ at time $k$, the reward function is designed as the sum of three terms. The first term $(r_g)^i_k$ is designed to encourage the agent for reaching the goal state,
\begin{align}
\label{eq:goalreward}
(r_g)^i_k = \begin{cases} r_{arrival} &\text{ if } ||\s^i_{k+1} - \tilde{\s}^i || < d_{arrival} \\
0 &\text{ otherwise}
\end{cases},
\end{align}
where $d_{arrival}$ represents a small, positive threshold to determine whether agent $i$ has reached the goal state $\tilde{\s}^i$, and $r_{arrival}$ is a positive reward.
The second term $(r_p)_k^i$ is designed to award the agent approaching the goal state. 
\begin{align}
(r_p)^i_k = - \frac{||\s^i_{k+1} - \tilde{\s}^i||}{||\s^i_0 - \tilde{\s}^i||}.
\end{align}
As the agent gets closer to the goal state, the reward becomes larger.
The last term $(r_c)^i_k$ is designed to penalize collisions with other agents,
\begin{align}
(r_c)^i_k = \begin{cases} r_{collision} \  &\text{if } ||\p^i_{k+1} - \p^j_{k+1} || < 2r, \ \forall j \neq i \\
0 &\text{ otherwise}
\end{cases},
\end{align}
where $r_{collision}$ is a negative penalty for collisions. 
% To award actions for reaching the goal state and penalize actions for collisions, we usually assign $r_{arrival}$ a large positive number and $r_{collision}$ a small negative number.

\textit{\textbf{Training Process:}} Since all agents are homogeneous, it is natural to assume that both agents follow the same policy. Therefore, when updating the neural network, the difference of parameters $d \theta$ comes from the observation, action and reward trajectories of both agents. This is different from standard RL, which assumes there exists only one decision maker. This modification could be applied to nearly any deep RL framework. In our experiment, we use Evolution Strategy (ES) as our training method.

\subsection{Multi-agent Collision Avoidance with Improved Safety}\label{sub:multi}
For multi-agent collision avoidance problem, each agent has more than one neighbor agent. Given the trained policy from two agents collision avoidance problem, each nearby agent would introduce an action for collision avoidance. To avoid collisions in multi-agent problems, multiple actions are combined into one action. Then we introduce linear safety constraints from the ORCA and handle the multi-agent collision avoidance problem by solving a simple convex optimization. 

\textit{\textbf{Action Combination:}} For one agent $i$, each neighbor agent $j$ results in an action from the trained policy, $\a^{i,j}_k \triangleq \underline{\pi}(\o^{i,j}_k, \underline{\theta})$, $j\neq i$ and $i,j = 1,\ldots, M$. Intuitively, the actions introduced by closer neighbors and approaching neighbors are more important. Therefore, we combine all actions with distance and velocity-based weights:
\begin{align}
\label{eq:combine}
\a^{i, comb}_k &= \frac{\sum_{j=1, j \neq i}^{M} \w_{i,j} \a^{i,j}_k}{\sum_{j=1, j \neq i}^{M} \w_{i,j}} \\
\mathbf{w}_{i,j} & = \frac{e^{-\alpha D_{i,j}}}{D_{i,j}}.
\end{align} 
Here $\w_{i,j}$ represents the weight determined by the pseudo-distance $D_{i,j}$, and $\alpha$ is a scaling parameter. Our design of weight $\w_{i,j}$ is inspired by the artificial potential function proposed by \cite{wolf2008artificial}. The weight $\w_{i,j}$ becomes infinite when $D_{i,j} = 0$. The pseudo-distance $D_{i,j}$ is determined by the relative position of agent $j$ to agent $i$ and velocity of agent $j$. If agent $i$ is behind the neighbor agent $j$, $D_{i,j}$ is the Euclidean distance, otherwise it is defined as adjusted distance that scaled along the velocity of agent $j$:
\begin{align}
    D_{i,j} &= 
    \begin{cases}
    ||\p^{j,i}|| & \text{ if } (\p^{j,i})^T \ve^j < 0 \\
    \sqrt{\frac{||\p^{j,i} \times \ve^j||^2 + ||\gamma_{j} \cdot (\p^{j,i})^T \ve^j||^2}{||\ve^j||^2}} & \text{ if } (\p^{j,i})^T \ve^j \geq 0 
    \end{cases} \\
    \gamma_{j} &= e^{-\beta ||\ve^j||}.
\end{align}
Here $\p^{j,i}$ represents the relative position of agent $j$ to $i$, and $\gamma_j$ represents the scaling factor that defined by parameter $\beta$ and $\ve^j$, i.e. the velocity of agent $j$. Such a design of weights makes agent $i$ to give priority to agent $j$ that is approaching, and the weight increases as agent $j$ speeds up. This weighted combination approach is consistent with human intuition. As human drivers, we would pay more attention to cars in the vicinity and approaching. Given the combined action, it is possible that the agent $i$ would collide with nearby agents in complex multi-agent cases. To guarantee safety, we introduce the well-known ORCA constraints and address the multi-agent collision avoidance problem by solving a simple convex optimization.

\textit{\textbf{A Brief Review of ORCA: }}Before introducing the convex optimization, we briefly review some key concepts in the optimal reciprocal collision avoidance (ORCA) algorithm. The ORCA algorithm is a distributed collision avoidance algorithm for continuous time system with single integrators dynamics.
% , which assumes that the velocity of each agent could be directly controlled and changed instantaneously.
Similarly to our assumption, ORCA assumes that each robot could be approximated by a disc with different radius $r_i$. Each agent takes action independently by observing the velocities $\ve^i(t)$, positions $\p^i(t)$ and radius $r_i$ of other agents. Since ORCA is designed for agents with continuous time domain, with a little abuse of notation, here we use $\ve^i(t)$ and $\p^i(t)$ to represent velocity and position of agent $i$ at continuous time $t$. We first consider a two agents collision avoidance problem with a continuous time horizon $T$. For two agents $i$ and $j$ at time $t$, ORCA returns a pair of linear constraints for velocities of the two agents independently:
\begin{equation}
    \begin{cases}
    (\a^{i,j})^{T}\ve^i(t + \tau) + b^{i,j} \geq 0 \ \tau \in [0, T]\\
    (\a^{j,i})^{T}\ve^j(t + \tau) + b^{j,i} \geq 0 \ \tau \in [0, T]
    \end{cases}.
\end{equation}
Here $\a^{i,j}$, $\a^{j,i}\in \R^d$ and $b^{i,j}$, $b^{j,i} \in \R $ are determined by the current positions $\p^i(t)$, $\p^j(t)$, velocities $\ve^i(t)$, $\ve^j(t)$ and radius $r_i$, $r_j$, which are known to both agents. If both constraints are satisfied, there would be no collision between agents $i$ and $j$ within time horizon $T$. When ORCA is extended to the multi-agent problem, there would be multiple linear constraints for each agent $i$,
\begin{equation} 
\label{ieq:orcac}
A_i \ve^i(t + \tau) + \mathbf{b}_i \geq 0, \ \tau \in [0, T],
\end{equation}
where $A_i \in \R^{K\times d} $ and $\mathbf{b}_i \in \R^K$, $K$ is the number of nearby agents to be considered. Note that $K$ may be smaller than $M-1$, which means we do not need to consider all moving agents for collision avoidance of agent $i$. Given the maximal velocity $v^{j,max}$ of agent $j$, if distance of agent $i$ and $j$ is greater than $(v^{i,max} + v^{j,max})T$, they would not collide within time horizon $T$. With properly selected $K$, inequality (\ref{ieq:orcac}) provides a safety guarantee of each agent $i$ in velocity space.
% The agent $i$ selects a new velocity $\ve^{i,new}$ by solving the following optimization problem,
% \begin{subequations}
% \begin{align}
%     \ve^{i,new} &= \argmin_{\ve} ||\ve - \ve^{i, pref}|| \\
%     \text{s.t. } &A_i \ve + \mathbf{b}_i \geq 0 \\
%     & ||\ve|| < v^{i,max} \label{cons:maxv}
% \end{align}
% \end{subequations}
% Here $\ve^{i, pref}$ is the preferred velocity from high-level planning and $v^{i,max}$ is the maximum speed of agent $i$.

\textit{\textbf{Convex Optimization with Safety Constraints:}} To avoid collisions for general nonlinear agents, we introduce the ORCA constraints~(\ref{ieq:orcac}) and transform them from velocity space to action space. Different from ORCA, our approach is designed for discrete time systems. We first adapt~(\ref{ieq:orcac}) for discrete time systems by letting $T \triangleq N \triangle t$ and $t \triangleq k \triangle t$. Here $N$ is a positive integer representing discrete time horizon, and $k$ is a non-negative integer corresponding to current time $t$, and $\triangle t$ is the discretization time step. The ORCA constraints could be written as 
\begin{equation}
\label{ieq:orcad}
A_i \ve^i_{k+d} + \mathbf{b}_i \geq 0, \ d = 1, \ldots, N.
\end{equation}
In practice, we update $A_i \in \R^{K\times d}$ and $\mathbf{b}_i \in \R^{K}$ that defines the ORCA constraints ($\ref{ieq:orcad}$) at each time step, thus it suffices to choose $d = 1$, leading to
\begin{equation}
    \label{ieq:orcafinal}
    A_i \ve^i_{k+1} + \mathbf{b}_i \geq 0.
\end{equation}
Here $A_i $ and $\mathbf{b}_i$ is determined by positions, velocities (and shared radius $r$) of agent $i$ and the closet $K$ neighbor agents at time $k$. Safety is guaranteed if (\ref{ieq:orcafinal}) holds at each time step. To control the agent $i$ for collision avoidance, we need to change the linear constraints from velocity space to action space. We assume the relationship between current action $\a^i_k$ and velocity at next time step $\ve^i_{k+1}$ is linear or could be approximated by a linear mapping. For a general nonlinear system, we could linearize the system around the current state and last action. This relationship is approximated by 
\begin{equation}
\label{eq:linear}
\ve^i_{k+1} \approx A_v \a^i_k + \mathbf{b}_v.
\end{equation}
By combing~(\ref{ieq:orcafinal}) and~(\ref{eq:linear}), we could obtain a safety constraints for the action $\a^i_k$,
\begin{equation}
\label{eq:cons}
A_i (A_v \a^i_k + \mathbf{b}_v) + \mathbf{b}_i \geq 0.
\end{equation}
If inequality~(\ref{eq:cons}) holds at each time step and the relationship between action $\a^i_k$ and the velocity $\ve^i_{k+1}$ is exactly linear, it is strictly guaranteed that the agent $i$ would never collide with any other agents~\cite{van2011reciprocal}. Rigorously speaking, the linearization for nonlinear system can introduce small errors. However, in practice this can be worked around by slightly enlarging the radius $r$ in our approach.   

The combined action from RL~(\ref{eq:combine}) provides preferred action that would drive the robot to reach the goal state, and the linear constraints from~(\ref{eq:cons}) provides safety constraints for the agent. To reach the goal while avoiding collisions, agent $i$ selects the action that is the closest to $\a^{i,comb}_k$ while satisfying safety constraints~(\ref{eq:cons}) and physical constraints:
\begin{subequations}
\label{eq:mca}
\begin{align}
\a^i_k = & \argmin_{\a} ||\a - \a^{i,comb}_k||^2 \label{eq:cost} \\
\text{s.t. } &  A_i (A_v \a + \mathbf{b}_v) + \mathbf{b}_i \geq 0 \label{cons:orca} \\
& \a \in \A \label{cons:dyna}.
\end{align}
\end{subequations}
Here~(\ref{eq:cost}) is a quadratic cost function,~(\ref{cons:orca}) are linear safety constraints from ORCA, and~(\ref{cons:dyna}) is the convex physical constraints. The optimization problem~(\ref{eq:mca}) is a simple convex optimization and can be solved efficiently. 
% Note that given the selected action $\a^i_k $ by solving~(\ref{eq:mca}) at each time step, the agent $i$ would never collide with other agents and safety is guaranteed.

To sum up, given the trained policy $\underline{\pi}(\cdot, \underline{\theta})$ from the two agents collision avoidance problem, we propose a decentralized multi-agent collision avoidance algorithm by solving a simple convex optimization with safety constraints from ORCA. If the relationship between action and velocity is linear, the safety constraints provide strict safety guarantee. The overall approach is summarized in Algorithm~\ref{alg:mca}. In the next section, we demonstrate our algorithm via different challenging experiments.

\begin{algorithm}[t]
 	\caption{Reciprocal Collision Avoidance for General Nonlinear Agents using Reinforcement Learning} \label{alg:mca} 
 	\begin{algorithmic}[1]
 		\State	{\bf Input:}  The trained policy $\underline{\pi}(\cdot, \underline{\theta})$ from the two agents collision avoidance problem, the initial state $\s^i_{initial}$, and the goal state of each agent  $\tilde{\s}^i$, $i = 1, \ldots, M$
 		\State Initialization: $\s^i_0 \leftarrow \s^i_{initial}$, $k \leftarrow 0$
 		\While{not all agents reach goal states}
 		\For{ $i=1, \ldots, M$}
 		\For{ $j=1, \ldots, M$ and $j \neq i$}
 		\State $\a^{i,j}_k \leftarrow \underline{\pi}(\o^{i,j}_k, \underline{\theta})$
 		\EndFor
 		\State Combine actions, $\a^{i,comb}_k$ via~(\ref{eq:combine})
%  		\State Calculate safety constraints, $A_i$, $\mathbf{b}_i$, from ORCA
 		\State Select action $\a^i_k$ by solving optimization ($\ref{eq:mca}$)
 		\State Apply $\a^i_k$ to agent $i$
        \EndFor
        \State $k \leftarrow k + 1$
        \EndWhile
 	\end{algorithmic}
 \end{algorithm} 

\section{Experiments and Results}
We illustrate the performance of our proposed method through several multi-agent interactive scenarios. 

\subsection{Simulation Setup}
\begin{figure}[b]
    \centering
    \includegraphics[trim={0cm 0cm 0cm 0cm},clip,width=0.4\textwidth]{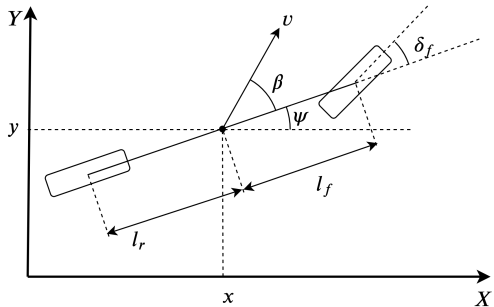}
    \caption{Kinematic Bicycle Model}
    \label{fig:bicycle}
\end{figure}
Throughout this section, we adopt the nonlinear kinematic bicycle model~\cite{rajamani2011vehicle} operating in a two-dimensional Euclidean space. 
As illustrated in Fig.~\ref{fig:bicycle}, the dynamics is defined as
\begin{subequations}
\label{eq:bicycle}
\begin{align}
    \dot{x} &= v \cos(\psi + \beta) \\
    \dot{y} &= v \sin(\psi + \beta) \\
    \dot{\psi} &= \frac{v}{l_r} \sin(\beta) \\
    \dot{v} &= a \\
    \beta &= \tan^{-1}\left(\frac{l_r}{l_f + l_r} \tan{\delta_f} \right) 
\end{align}
\end{subequations}
where $x, y$ are the coordinates of the centroid, $\psi$ is the inertial heading, $v$ is the speed of vehicle, $\beta$ is angle of velocity with respect to longitudinal axis of the car, $l_f$ and $l_r$ represents the distance from the centroid to the front and rear axles. Actions consist of front steering angle $\delta_f$ and acceleration $a$. 

Unlike other RL-based approaches~\cite{chen2017decentralized,chen2017socially,everett2018motion,fan2018fully,long2018towards} that assume that velocity of each agent could be directly controlled, our approach is directly applicable to the above nonlinear model with acceleration $a$ and steering angle $\delta_f$ as control. As far as we know, this is the first RL-based multi-agent collision avoidance algorithm that directly works for kinematic bicycle model. As for ORCA-based approach,~\cite{alonso2012reciprocal} is designed for bicycle model by fixing a tracking controller and transferring the control to velocity space. But like many other ORCA-based approaches, it requires a high-level planning module to provide preferred velocity, which would greatly influence the performance.

% In simulation, we assume that each agent observes positions and velocities of up to four nearest surrounding agents for ORCA constraints, that is, $K = 4$. 
To reduce redundancy and simplify training, we use a local coordinate for each agent with $x$ axis pointing towards the inertial heading of the vehicle in simulation. The observation of agent $i$ at time $k$ is represented by concatenating the goal position, velocity of agent $i$, positions and velocities of nearby agents in the local coordinate. The dynamics~(\ref{eq:bicycle}) is discretized in time and all simulations operate with discretization time step of 0.05 $s$ (20 Hz).

\subsection{Policy Training with RL}
The policy for two-agent collision avoidance is trained following the described procedure in Section~\ref{sub:two}. We use a two-layer neural network with 16 neurons at each layer. The network takes the observation of size 8 and predicts the control action of size 2. Compared with other RL-based approach, this is a significantly lightweight design of network architecture with a total of only 450 training parameters. On an eight-core i7-7700K CPU 4.20GHz machine, the empirically optimal policy is obtained within 20 minutes of wall-clock time training. The policy is then combined with the ORCA constraints to handle multi-agent interactions. It is worth noting that all simulation results in~\ref{simuresults} comes from the same trained policy.

\subsection{Simulation Results}\label{simuresults}

\begin{figure}
\centering
\includegraphics[trim={7cm 5cm 7cm 8cm},clip,width=0.5\textwidth]{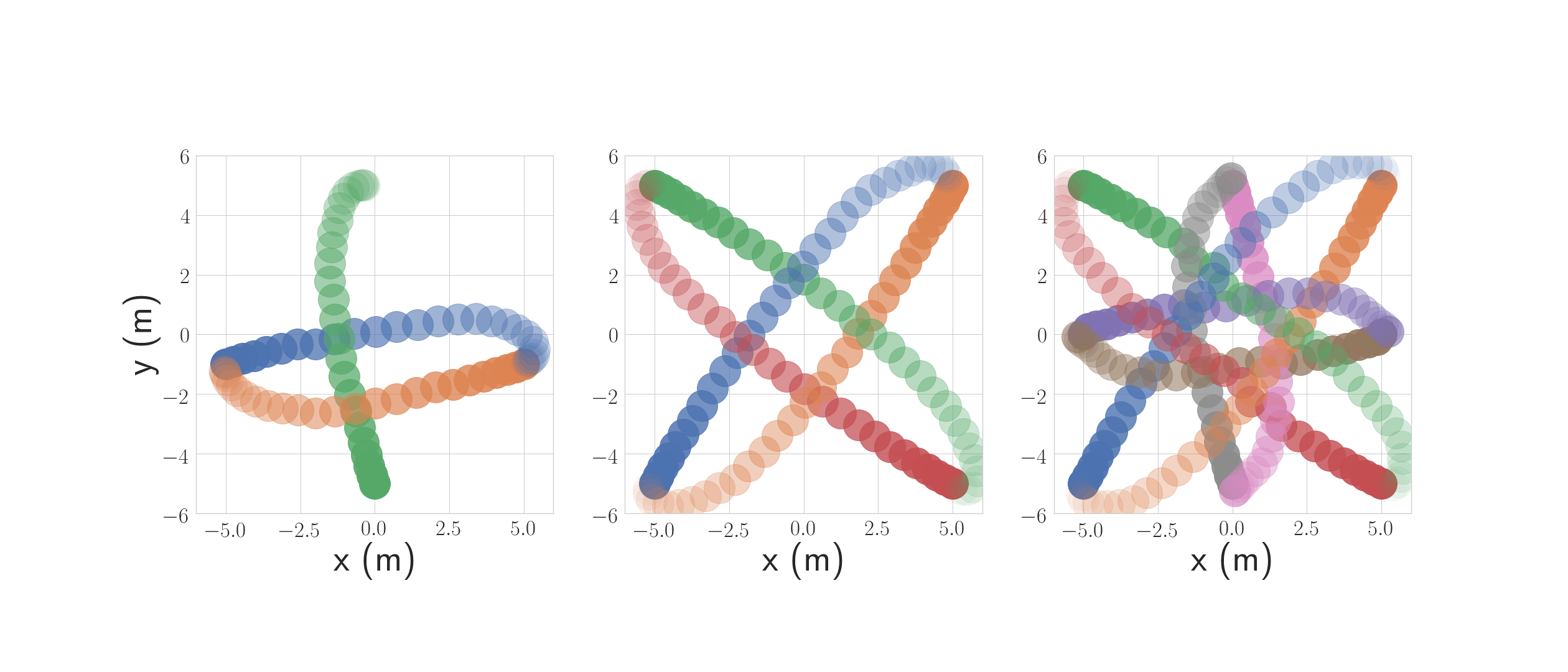}
\caption{Trajectories of three, four and eight agents}
\label{fig:348}
\end{figure}

\begin{figure}[t]
    \centering
    \includegraphics[trim={10cm 1cm 8cm 6cm},clip,width=0.5\textwidth]{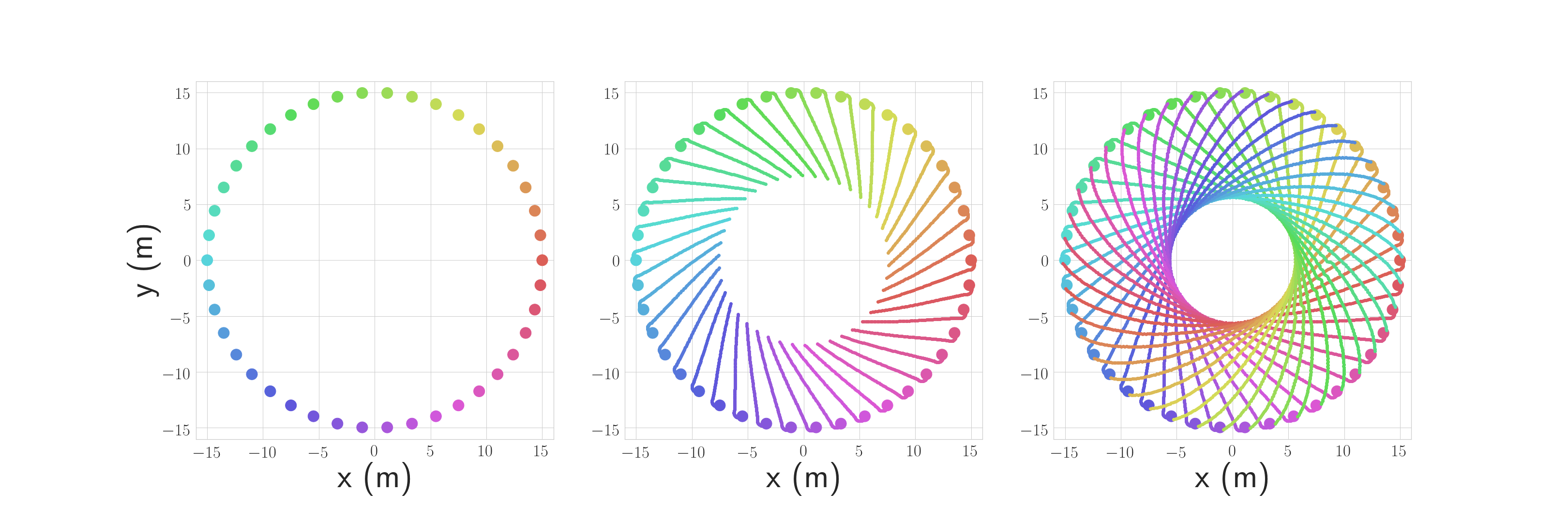}
    \caption{Trajectories of 42 agents. The left figure: the initial position of each agent, different agents are marked by dots with different color. The middle figure: the trajectory in the middle of the process. The right figure: complete trajectories.}
    \label{fig:circle}
\end{figure}

\begin{figure}[t]
    \centering
    \includegraphics[trim={10cm 1cm 8cm 6cm},clip,width=0.5\textwidth]{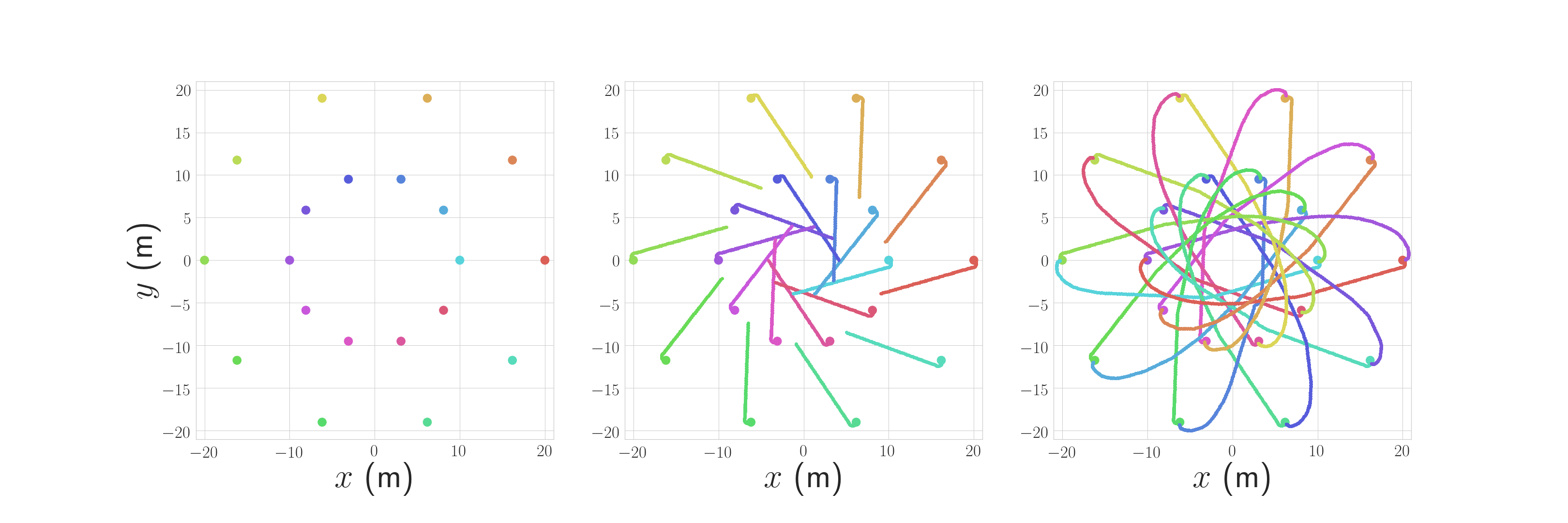}
    \caption{Trajectories of 20 agents. The left figure: the initial position of each agent is denoted by solid circle with different colors. The middle figure: the trajectory in the middle of the process. The right figure: complete trajectories.}
    \label{fig:flower}
\end{figure}

\begin{figure}[t]
\begin{subfigure}{.52\textwidth}
  \centering
  % include first image
  \includegraphics[trim={12cm 1cm 8cm 4.8cm},clip,width=1\textwidth]{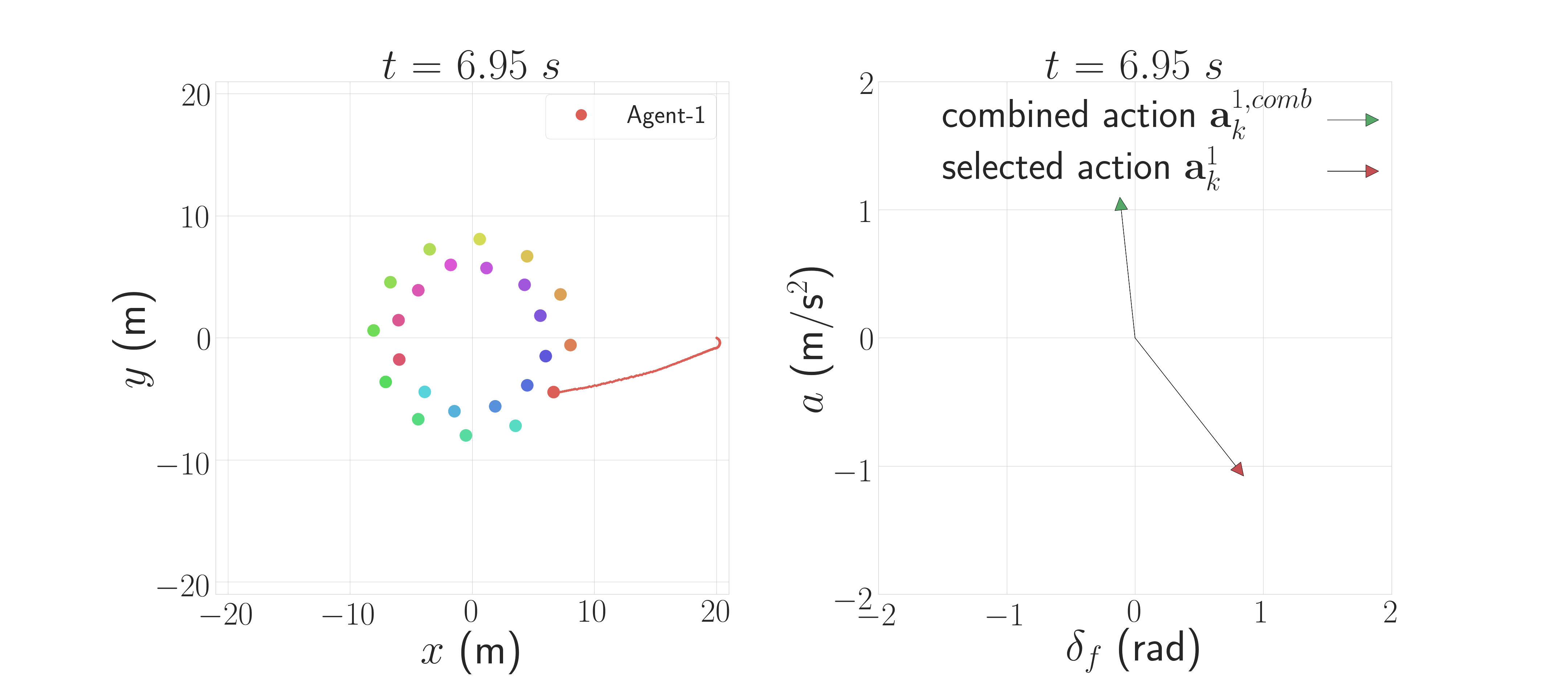}  
  \caption{Positions, trajectory, actions when $t=6.95$ $s$ }
  \label{fig:sub-first}
\end{subfigure}
\begin{subfigure}{.52\textwidth}
  \centering
  % include second image
  \includegraphics[trim={12cm 1cm 8cm 1cm},clip,width=1\textwidth]{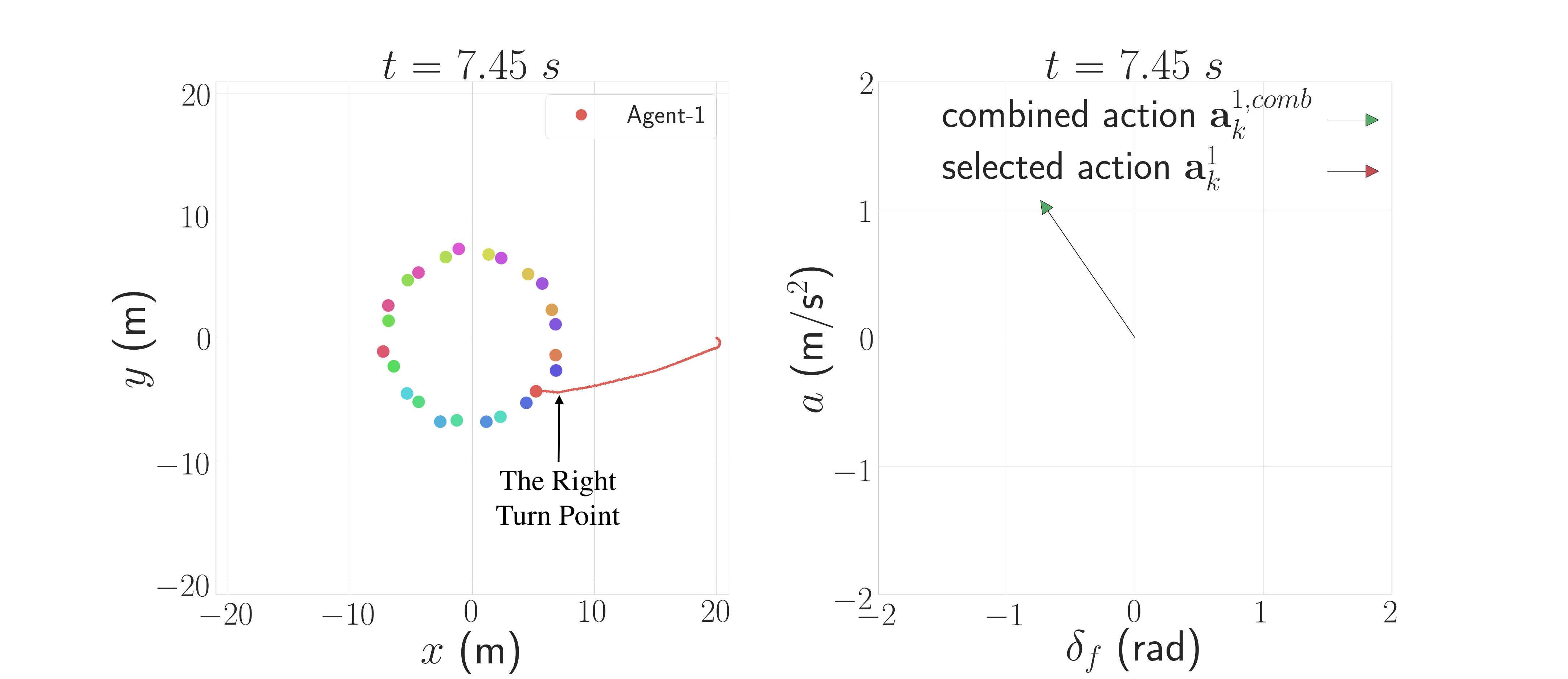}  
  \caption{Positions, trajectory, actions when $t=7.45$ $s$ }
  \label{fig:sub-second}
\end{subfigure}
\begin{subfigure}{.52\textwidth}
  \centering
%   % include second image
%   \includegraphics[trim={12cm 2cm 8cm 1cm},clip,width=1\textwidth]{images/0_163.png}  
%   \caption{Positions, trajectory, actions at $t=8.15$ $s$ }
%   \label{fig:sub-third}
\end{subfigure}
\caption{Illustration of the importance of ORCA constraints through a 24 agents example. The left column shows positions of all agents and trajectories of agent $1$. The right column shows combined actions and selected actions of agent $1$.}
\label{fig:flower_action}
\vspace{-0.1in}
\end{figure}

% $24$ agents test case. The left column shows positions of 25 agents, which is marked by dots with different color, and trajectory of agent $1$, which is marked by red solid lines. The right column shows the combined actions of agent $1$ from~(\ref{eq:combine}), which are denoted by green arrows, and selected actions by solving~(\ref{eq:mca}) with ORCA constraints, which are denoted by red arrows. When two actions coincide at the same time, only green arrow would be displayed. Positions, trajectories and actions at $t = 6.95$ $s$, $7.45$ $s$ are shown in (a) and (b) respectively.

% \begin{figure*}
% \begin{multicols}{2}
%     \includegraphics[trim=6cm 2cm 10cm 2cm, clip,width=0.48\textwidth]{images/0_139.png}\par 
%     \includegraphics[trim=6cm 2cm 10cm 2cm, clip,width=0.48\textwidth]{images/0_143.png}\par 
%     \end{multicols}
% \begin{multicols}{2}
%     \includegraphics[trim=6cm 2cm 10cm 2cm, clip,width=0.48\textwidth]{images/0_145.png}\par
%     \includegraphics[trim=6cm 2cm 10cm 2cm, clip,width=0.48\textwidth]{images/0_170.png}\par
% \end{multicols}
% \caption{Trajectory of one selected agent with ORCA constrained actions at selected time steps.}
% \label{fig:flower_action}
% \end{figure*}

As shown in Fig.~\ref{fig:348}, we first test the proposed method in simple scenarios with up to eight agents. The trajectory of each agent is represented by circles with the same color, and the color gradually fades as the agent moves. The goal position of each agent is symmetric with the initial position about the origin. 
It is worth emphasizing that the three-agent interactive scene is inspired by a well-know deadlock scenario of the classic ORCA method~\cite{godoy2018alan}. 
The symmetric setup of scenarios with four and eight agents are also easy to result in conflicts of actions. Surprisingly, our proposed approach successfully generates a set of smooth trajectories reaching the target position for all the three scenarios without encountering any collision, congestion or deadlock.

% The target position is symmetric with the initial position about the origin
% We first test our approach with three, four and eight agents to demonstrate that our approach could easily avoid deadlock. As shown in Fig.~\ref{}, from left to right, there are three, four and eight agents in each plot, the goal position is symmetric with the initial position about the origin. The trajectory of each agent is represented by circles with the same color, and the color gradually fades as the agent moves. As showed by \cite{godoy2018alan}, such tasks is not easy for many deterministic algorithm, since the symmetric initial state could easily result in deadlock, deadlock detection and heuristics are necessary to alleviate the problem. However, trajectories generated by our approach is very smooth and there is no collision, congestion or deadlock. Each agent could avoid each other very efficiently.

We now demonstrate our approach with a more complex task. As shown in Fig.~\ref{fig:circle}, we have 42 agents evenly distributed on a circle, with initial velocities pointing to the adjacent neighbor of each agent clockwise. The goal position of each agent is symmetric with the initial position about the origin. Trajectories of different agents are marked by different colors. It is clear from the figure that a set of safe and smooth trajectories are generated with each agent first turning right to have the heading more aligned with the target position, then passing the origin from the left side with ORCA-constrained actions for collision avoidance.

% We could find that each agent would turn right first to make the heading closer to its goal position, then passes the origin from the left side. All trajectories are of the same shape. Unlike many methods need heuristics to handle deadlock, all trajectories from our approach are safe and smooth.

% \begin{figure}[ht]
%     \centering
%     \subfigure[Time ]{
%     \includegraphics[0.1\linewidth]{images/0_139.png}
%     \caption{text}
%     }
% \end{figure}

The next example is to demonstrate the capability of the proposed method in a customized task that requires more sophisticated interactions with surrounding agents. As shown in Fig.~\ref{fig:flower}, initial positions are marked in the left figure as dots with different colors. We have 20 agents that are evenly distributed on two co-centric circles. For agents located on the large circle, the goal positions are located on the small circle that aligns with the start position through the center, and vice versa. Such a task is more difficult than previous ones. In order to reach the goal position, each agent is expected to have conflicts with agents initialized from both circles. However, as shown in Fig.~\ref{fig:flower}, trajectories generated by our approach is smooth and safe.

Furthermore, recall the discussion in Section~\ref{sub:multi} where the combination of actions from trained policy could also drive the controlled agent to the unsafe state. From Fig.~\ref{fig:flower_action}, we select one agent and further illustrates {\em how the ORCA constraints alter the selected actions to enhance safety}. As shown in Fig.~(\ref{fig:sub-first}), when $t = 6.95$ $s$, the combined action of agent $1$ would drive the robot forward to its left. However, this would cause a collision with the robot (marked by dark blue) to its front left. To avoid collision, the ORCA constraints revise the robot to a right turn with deceleration. We can observe an obvious twist to the right from the position trajectory in the left figure of Fig.(\ref{fig:sub-second}), which is marked by a black arrow. The example clearly indicates the importance of having ORCA constraints in the proposed approach to avoid collisions.

% are triggered and the selected action is to decelerate and turn right. From Fig.~(\ref{fig:sub-second}) and Fig.~(\ref{fig:sub-third}) we could find the agent passes by the robot marked by dark blue from the right side by slightly turning right. The position making the right turn is marked by a black arrow in the left figure of Fig.(~\ref{fig:sub-second}).
% Provide the combined actions at $6.95$ $s$, $7.45$ $s$ and $8.15$ $s$, a set of sel optimal actions are derived to help avoid collisions with the blue colored agent. 

\section{Conclusion}
In this paper, we propose a decentralized collision avoidance algorithm for general nonlinear systems by combing the RL and ORCA. The proposed method consists of two stages: first train a two-agent collision avoidance policy using RL; then compute the overall multi-agent collision avoidance action by properly combining multiple avoidance actions while respecting safety constraints induced by ORCA. The overall training process is much faster than other RL-based collision avoidance algorithms. The introduction of ORCA significantly improves system safety. We demonstrate the proposed method through several collision avoidance problems with kinematic bicycle agent models. In the future, we would like to apply the proposed approach to systems with more complex dynamics which has larger state space.

\newpage
\bibliography{references.bib}
\bibliographystyle{IEEEtran}
\end{document}